\documentclass[letterpaper, 10 pt, conference]{ieeeconf}  

\usepackage{graphicx}
\usepackage{multirow}
\usepackage{amsmath}
\usepackage{amssymb}
\usepackage{array}
\usepackage{subcaption}
\usepackage[dvipsnames]{xcolor}
\usepackage{color,soul}
\usepackage{url}

\IEEEoverridecommandlockouts                              

\title{\LARGE \bf
Robust Unmanned Surface Vehicle Navigation with Distributional Reinforcement Learning
}

\author{Xi Lin, John McConnell and Brendan Englot
\thanks{Xi Lin, John McConnell and Brendan Englot are with the Department of Mechanical Engineering, Stevens Institute of Technology, 1 Castle Point Terrace, Hoboken, NJ 07030, USA,
{\tt\small \{xlin26,jmcconn1,benglot\}@stevens.edu}}%
}

\begin{document}

\maketitle
\thispagestyle{empty}
\pagestyle{empty}

\begin{abstract}
Autonomous navigation of Unmanned Surface Vehicles (USV) in marine environments with current flows is challenging, and few prior works have addressed the sensor-based navigation problem in such environments under no prior knowledge of the current flow and obstacles.
We propose a Distributional Reinforcement Learning (RL) based local path planner that learns return distributions which capture the uncertainty of action outcomes, and an adaptive algorithm that automatically tunes the level of sensitivity 
to the risk in the environment.
The proposed planner achieves a more stable learning performance and converges to safer policies than a traditional RL based planner.
Computational experiments demonstrate that comparing to a traditional RL based planner and classical local planning methods such as Artificial Potential Fields and the Bug Algorithm, the proposed planner is robust against environmental flows, and is able to plan trajectories that are superior in safety, time and energy consumption. 
\end{abstract}

\section{Introduction}
The operation of autonomous vehicles in marine environments is sensitive to 
currents \cite{lolla2015path}, and navigating safely and efficiently under the influence of current flow is challenging.
In recent years, several methods \cite{song2017multi,guo2020global,meng2022anisotropic} have been proposed to deal with the global path planning problem for Unmanned Surface Vehicles (USV) given knowledge of the current flow field.
However, few prior works have focused on the local path planning problem for USVs in environments with unknown current flows.   
In this work, inspired by recent USV competitions \cite{robotx} and by the challenges of navigating in river rapids, we consider the sensor-based local navigation problem for USVs in simulated marine environments with no prior knowledge of the current flow and obstacles.

Reinforcement Learning (RL) shows the ability to acquire high-performance policies to operate in unseen environments by learning through experiences of interaction with training environments given no prior information \cite{sutton2018reinforcement}.
In the last decade, Deep Reinforcement Learning (DRL) combining RL with deep neural network architectures has been widely used to solve practical problems with high-dimensional sensory inputs, which can be efficiently executed via neural network inference.
Compared to a traditional DRL method that only learns the expected return, a Distributional Reinforcement Learning (Distributional RL) method is shown to provide a more stable learning behavior in environments with high uncertainty as it learns return distributions \cite{bellemare2017distributional}.
In addition, risk measures can be applied to adjust the level of 
sensitivity to aleatoric
uncertainty in learned distributions and enhance the safety performance of a Distributional RL agent \cite{dabney2018implicit}.

\begin{figure}
    \centering
    \includegraphics[width=\linewidth]{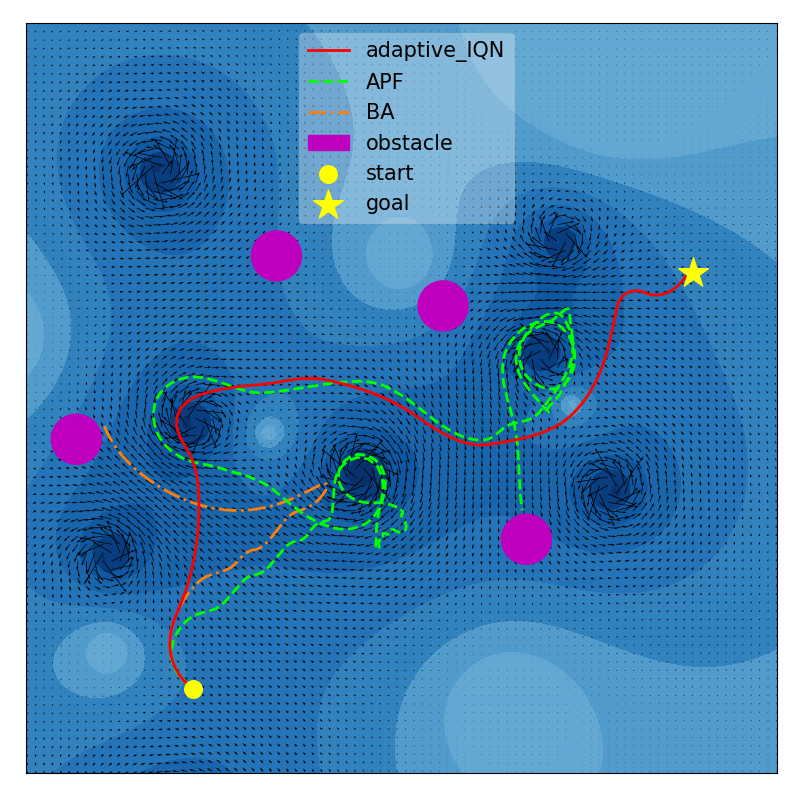}
    \caption{\textbf{Comparison of planned trajectories.} Planned trajectories of the proposed Distributional RL based path planner and classical planners in a simulated marine environment with no prior knowledge of the current flow field and obstacles, where darker color indicates faster current flows. 
    Arrows show the velocity of the current in specific locations.
    }
    \label{fig:demonstration}
    \vspace{-3mm}
\end{figure}

We propose a Distributional RL based local path planner to address USV sensor-based navigation tasks under unknown obstacles and current flows, as well as an algorithm that adjusts the level of sensitivity 
towards collision risk and further improves the task success rate. 
The proposed method exhibits a more stable learning performance and converges to safer policies compared to a traditional DRL-based local planner.  
We compare the proposed method's performance to classical methods including Artificial Potential Fields \cite{song2020path,sang2021hybrid} and the Bug Algorithm \cite{wilson2018adaptive,lyridis2021improved}, which have been applied to solve the USV local path planning problem in stable marine environments with minimal current disturbances. 
It is shown in Figure \ref{fig:demonstration} that the performances of Artificial Potential fields and the Bug Algorithm are vulnerable to unknown current flows, and result in unsafe, zig-zag trajectories. 
On the contrary, the proposed Distributional RL planner is able to plan a much smoother trajectory under current disturbances while keeping a safe distance from detected obstacles, thus achieving superior performance in safety, time and energy consumption.
Our contributions are summarized as follows:
\begin{itemize}
    \item To our knowledge, the first Distributional RL based path planner for USV sensor-based navigation in environments with unknown current flows and obstacles.  
    \item Simulated experiments that show superior performance in safety, time and energy consumption vs. traditional RL and classical reactive planning algorithms. 
    \item The software implementation of our approach and a simulation environment for studying decision making in USV navigation (amidst unknown currents and obstacles) have been made freely available at \url{https://github.com/RobustFieldAutonomyLab/Distributional_RL_Navigation}.
\end{itemize}

The rest of this paper is organized as follows: Section \ref{sec:related works} introduces related work, highlighting relevant local path planning methods; Section \ref{sec:problem formulation} describes the RL problem formulation; Section \ref{sec:methodology} introduces the methodology of this work; Section \ref{sec:experiments} introduces baseline approaches and discusses experimental results; Section \ref{sec:conclusion} concludes the paper. 

\section{Related Works}
\label{sec:related works}
Khatib \cite{khatib1986real} proposed to perform mobile robot navigation with an Artificial Potential Field (APF), which creates attractive force towards the goal and repulsive force to avoid obstacles.
However, the traditional APF method may fail when local minima are present, or the goal is close to obstacles, and solutions such as virtual local target \cite{li2012efficient}, deterministic annealing \cite{doria2013algorithm}, dynamic window \cite{sun2019smart}, and additional repulsive force \cite{weerakoon2015artificial,fan2020improved} were proposed to address this problem.
Some works focus on improving performance of the APF method in dynamic environments, by using a new potential function that considers the relative position and velocity of the robot with respect to the target and obstacles \cite{ge2002dynamic}, or evolutionary algorithms \cite{montiel2015optimal,orozco2019mobile}.

Bug Algorithms (BA) have been used for reactive mobile robot path planning since the 1980s \cite{mcguire2019comparative}.
Lumelsky et al. \cite{lumelsky1986dynamic} proposed the Bug1 and Bug2 algorithms, which use a point robot model and zero-range sensor for obstacle detection, and exhibit two behaviors: the robot moves in a straight line to the goal and follows the obstacle boundary as soon as it encounters an obstacle. 
The Bug2 algorithm has been applied to the local navigation problem on platforms such as wheeled robots \cite{zhu2010new}, quadrotors \cite{marino2016minimalistic} and USVs \cite{lyridis2021improved}. 
Alg1 and Alg2 \cite{sankaranarayanan1990new}, Rev1 and Rev2 \cite{horiuchi2001evaluation} improve the performance of bug algorithms by switching the obstacle following direction when the robot is in a previously visited location.
VisBug \cite{lumelsky1988paradigm,lumelsky1990incorporating} and TangentBug \cite{kamon1996new} use range sensors for obstacle detection and find shorter paths to the goal \cite{lyridis2021improved}.

Deep Reinforcement Learning (DRL) based planners have shown strong ability to navigate in unknown environments in recent years.
Tai et al. \cite{tai2017virtual} trained a DRL agent for navigation in unseen indoor environments without any collisions.
Zhang et al. \cite{zhang2018robot} and Josef et al. \cite{josef2020deep} showed robust performance of DRL planners on Unmanned Ground Vehicle (UGV) navigation in simulated unknown rough terrain where the surface normal could abruptly change. 
Cheng et al. \cite{cheng2018concise} proposed a DRL based path planner to control an underactuated USV under unknown environment dynamics. 
Xu et al. \cite{xu2020intelligent} developed a DRL agent for a USV to avoid dynamic obstacles such as boats.   

More recently, Distributional RL \cite{bellemare2017distributional} methods have been used to learn risk-aware navigation policies for enhanced safety.
Choi et al. \cite{choi2021risk} proposed a novel Distributional RL algorithm for indoor navigation tasks, which shows superior performance in safety relative to traditional DRL methods, as well as adjustable sensitivity 
towards collision risks.
Kamran et al. \cite{kamran2021minimizing} demonstrated that a Distributional RL based policy reduces driving time compared to traditional DRL methods, and requires much less safety interference than rule-based policies in autonomous driving scenarios.    
Liu et al. \cite{liu2022adaptive} proposed an algorithm that automatically adjusts the level of sensitivity toward risk of a Distributional RL planner for nano drone navigation.   

\section{Problem Formulation}
\label{sec:problem formulation}
We formulate the USV navigation problem as a Markov Decision Process $(\mathcal{S},\mathcal{A},\mathcal{P},R,\gamma)$, where $\mathcal{S}$ and $\mathcal{A}$ are the sets of states and actions. 
At each time step $t$, the agent receives an observation of the current state $s_t \in \mathcal{S}$, and selects an action $a_t \in \mathcal{A}$. 
This causes the 
agent to transition to the state $s_{t+1}\sim \mathcal{P}(\cdot|s_t,a_t)$ , and the agent receives an observation of $s_{t+1}$, as well as a reward $r_{t+1} = R(s_{t+1},a_{t+1})$.
A common way to find an optimal policy in RL is to maximize the expected future return $Q^\pi$.
\begin{equation}
  \label{eq:expected return}
  Q^\pi(s, a) = \mathbb{E}_\pi[\sum_{k=0}^\infty \gamma^k r_{t+k+1}| s_t =s, a_t=a]
\end{equation}
This is also known as the action-value function, which starts with a state-action pair $(s_t,a_t)$ and follows the policy $\pi$ thereafter.
Discount factor $\gamma\in [0,1)$ controls the effect of future rewards. 
\begin{equation}
    \label{eqn:Bellman}
    Q^\pi(s,a) = \mathbb{E}[R(s,a)] + \gamma\mathbb{E}[Q^\pi(s',a')]
\end{equation}
\begin{equation}
    \label{eqn:Bellman optimality operatoer}
    \mathcal{T}Q(s,a) := \mathbb{E}[R(s,a)] + \gamma\mathbb{E}[\max_{a'}Q(s',a')]
\end{equation}
The relation between $Q^\pi$ and its successors satisfies the Bellman equation (Eq. \eqref{eqn:Bellman}), where $s'\sim \mathcal{P}(\cdot|s,a)$, and $a'\sim \pi(\cdot|s')$. An optimal policy can be derived using the Bellman optimality operator (Eq. \eqref{eqn:Bellman optimality operatoer}).

\section{Methodology}
\label{sec:methodology}

\subsection{Traditional DRL}
We choose DQN \cite{mnih2015human} as the traditional DRL baseline, and use the implementation from the Stable-Baselines3 project \cite{stable-baselines3}. 
DQN parameterizes an approximate action-value function (Eq. \eqref{eq:expected return}) as $Q(s,a;\theta)$ with a neural network model, and learns an optimal policy by performing optimization on a loss function (Eq. \eqref{eq:DQN loss}) based on Temporal Difference (TD) error, where $(s,a,r,s')$ are samples from experiences.
\begin{equation}
    \label{eq:DQN loss}
    \mathcal{L}_{\text{DQN}} = \mathbb{E}[(r+\gamma\max_{a'}Q(s',a';\theta^-)-Q(s,a;\theta))^2]
\end{equation}
\subsection{Distributional RL}
Instead of the expected return (Eq. \eqref{eq:expected return}), Distributional RL algorithms  \cite{bellemare2017distributional} focus on the return distribution, which satisfies the distributional Bellman equation (Eq. \eqref{eqn:Distributional bellman}), where $Z^\pi(s,a)$ is a random variable that satisfies $Q^\pi(s,a) = \mathbb{E}[Z^\pi(s,a)]$. Similarly, the distributional Bellman optimality equation (Eq. \eqref{eqn:Distributional bellman optimality operator}) is used in this case.
\begin{equation}    \label{eqn:Distributional bellman}
    Z^\pi(s,a) \overset{D}{=} R(s,a) + \gamma Z^\pi(s',a') 
\end{equation}
\begin{equation}\label{eqn:Distributional bellman optimality operator}
    \mathcal{T}Z(s,a) :\overset{D}{=} R(s,a) + \gamma Z^\pi(s',\text{argmax}_{a'} \mathbb{E}[Z(s',a')]) 
\end{equation}
The proposed Distributional RL based path planner uses Implicit Quantile Networks (IQN) \cite{dabney2018implicit}, and we use the implementation from \cite{IQN_and_Extensions}.
IQN uses the quantile function $Z$, denoted as $Z_\tau := F_Z^{-1}(\tau)$, where $\tau\sim U([0,1])$, and a distortion risk measure $\beta:[0,1]\rightarrow[0,1]$ to define a distorted expectation.
\begin{equation}
    Q_\beta(s,a) = \mathbb{E}_{\tau\sim U([0,1])}[Z_{\beta(\tau)}(s,a)]
\end{equation}
Then the risk-sensitive policy can be expressed as $\pi_\beta(s) = \text{argmax}_a Q_\beta(s,a)$, and approximated by $K$ samples of $\Tilde{\tau}\sim U([0,1])$:
\begin{equation}
    \label{eq:approximate pi beta}
    \Tilde{\pi}_\beta(s) = \text{argmax}_a \frac{1}{K}\sum_{k=1}^K Z_{\beta(\Tilde{\tau}_k)}(s,a).
\end{equation}
The IQN loss function (Eq. \eqref{eq:IQN loss}) is constructed with the sampled temporal difference (TD) error (Eq. \eqref{eq:sampled TD}) and the quantile Huber loss (Eq. \eqref{eq:quantile Huber}).
\begin{equation}
    \label{eq:sampled TD}
    \delta^{\tau_i,\tau'_j} = r + \gamma Z_{\tau'}(s',\pi_\beta(s')) - Z_{\tau}(s,a)
\end{equation}
\begin{equation}
    \label{eq:quantile Huber}
    \begin{array}{c}
        \rho_\tau^\kappa(u) = |\tau-\delta_{\{u<0\}}|(\mathcal{L}_\kappa(u)/\kappa), \\[5pt]

        \text{where }\mathcal{L}_\kappa(u) = \left\{
            \begin{array}{lr}
                 \frac{1}{2}u^2, &\text{if}\ |u|\leq\kappa \\
                 \kappa(|u|-\frac{1}{2}\kappa), & \text{otherwise} 
            \end{array}
        \right.
    \end{array}
\end{equation}
\begin{equation}
    \label{eq:IQN loss}
    \mathcal{L}_{\text{IQN}} = \frac{1}{N'}\sum_{i=1}^N \sum_{j=1}^{N'} \rho_{\tau_i}^\kappa(\delta^{\tau_i,\tau'_j})
\end{equation}

\subsection{USV Navigation Environment}
We design a simulated marine environment with environmental flows and static obstacles, where the robot is required to navigate under flow disturbances and reach the goal without colliding with any obstacles.

The Rankine vortex model \cite{acheson1991elementary} (Eq. \eqref{eq:rankine}) is used to create the flows, where a rigid body rotation within the vortex core of radius $r_0$ is assumed, and $\Gamma$ is the circulation strength of the vortex.
\begin{equation}
    \label{eq:rankine}
    v_r = 0,\quad v_\theta(r) = \frac{\Gamma}{2\pi}\left\{
            \begin{array}{lr}
                 r/r_0^2, &\text{if}\ r\leq r_0 \\
                 1/r, &\text{if}\ r>r_0
            \end{array}
        \right.
\end{equation}
The angular velocity of the vortex core $\Omega=\Gamma / (2\pi r_0^2)$, and the linear velocity at the edge of vortex core $v_{\text{edge}}=\Omega r_0$.
The flow velocity at a specific location is approximated by superimposing the effects of nearby vortices.

We use a simplified USV model, the pose of which can be represented as $(x,y,\theta)$, where $(x,y)$ are the global Cartesian coordinates of the robot, and $\theta$ is its orientation.  
To consider the effects of current flow on robot motion, the kinematic model described in \cite{lolla2014time} is used, which treats the robot as a point particle, since its dimension is much smaller than the length of its trajectory and the scale of current flow.  
\begin{equation}
    \label{eq:pointrobot}
    \frac{d\mathbf{X}(t)}{dt} = \mathbf{V}(t) =  \mathbf{V}_C(\mathbf{X}(t)) +\mathbf{V}_S(t)
\end{equation}
In Eq. (\ref{eq:pointrobot}), $\mathbf{V}(t)$ is the total velocity of the robot, $\mathbf{V}_C(\mathbf{X}(t))$ is the current flow velocity at robot position $\mathbf{X}(t)$, and $\mathbf{V}_S(t)$ is the robot steering velocity.

\subsection{States and Observations}

The full state consists of  robot position $\mathbf{X}(t)$, robot steering velocity $\mathbf{V}_S(t)$, current flow field $\mathbf{V}_C(\mathbf{X})$, the positions and sizes of all obstacles, and the goal location.
In this work, the robot only receives a partial observation of the full state from onboard sensors, $O_t = (O_{\text{velocity}},O_{\text{goal}},O_{\text{LiDAR}})$, where $O_{\text{velocity}}$ is the robot's seafloor-relative velocity measured by Doppler Velocity
Log (DVL), $O_{\text{goal}}$ is the goal position in the robot frame computed with the aid of an inertial measurement unit (IMU), compass and GPS, and 
$O_{\text{LiDAR}}$ are LiDAR reflections indicating any obstacles ahead.
The LiDAR range is $d_0 = 10$ meters.
These assumptions are inspired by the class of USV required to successfully detect and avoid obstacles in the Maritime RobotX competition \cite{robotx2}.
An observation example is shown in Fig. \ref{fig:cvar_distributions}.

\subsection{Actions}
The action used to control the robot motion is the change in steering velocity $\mathbf{V}_S$.
We assume that the rates of change in both the magnitude and direction of $\mathbf{V}_S$ are constant across one control time step, hence an action is given in the form of $a_t = (a,w)$, where $a$ is the rate of change in velocity magnitude, and $w$ is the rate of change in velocity direction.
For convenience, $a$ and $w$ are referred to as linear acceleration and angular velocity.
We use $a \in \{-0.4,0.0,0.4\} \text{ m}/\text{s}^{2}$, and $w \in \{-0.52,0.0.0.52\} \text{ rad}/\text{s}$.
The forward speed is clipped to $[0,v_{max}]$. 

\subsection{Reward}
The reward function is designed to encourage the learning agent to move towards the goal while avoiding obstacles, which is shown in Equation \eqref{eqn:reward}.
\begin{equation}
    r_t = \left\{
        \begin{array}{cc}
            r_{\text{base,t}} + r_{\text{collision}}, & \text{if collide at } t \\
            r_{\text{base,t}}+r_{\text{goal}}, & \text{if reach goal at } t \\
            
            r_{\text{base,t}} = r_{\text{step}} + \alpha(d_{t-1}-d_{t}), &\text{otherwise}
        \end{array}
    \right.
    \label{eqn:reward}
\end{equation}
$d_t$ represents the distance between the robot and the goal at $t$. 
We use $r_{\text{step}} = -1.0$, $r_{\text{collision}} = -50.0$, $r_{\text{goal}} = 100.0$, and $\alpha = 1.0$ in our computational experiments.

\subsection{Network Architecture}
\begin{figure}
    \centering
    \includegraphics[width=\linewidth]{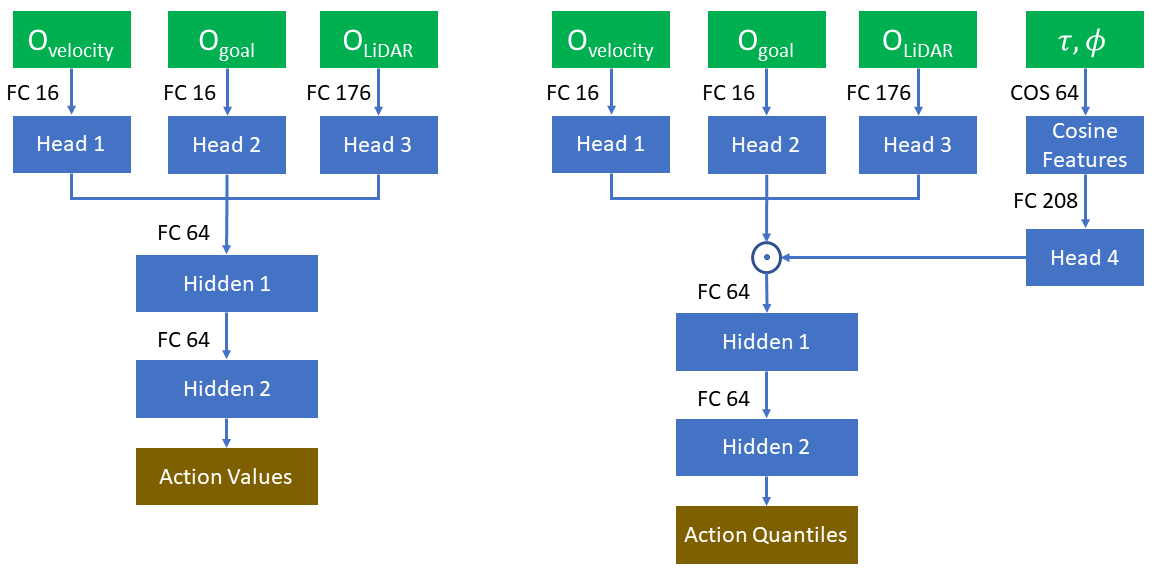}
    \caption{\textbf{Network models.} Architectures used by DQN (left) and IQN (right). FC, COS and $\odot$ stand for fully connected layer, cosine embedding layer and element-wise product.}
    \label{fig:networks}
\end{figure}

The network architectures used by our DQN and IQN agents are shown in Figure \ref{fig:networks}.
Observations from different sources are encoded separately as different features, which are then concatenated into a general one. Compared to the DQN network model, IQN employs an additional cosine function to embed the quantile input \cite{dabney2018implicit}, and we use conditional value-at-risk (CVaR, Eq. \eqref{eq:cvar}) as the distortion function. The resulting cosine features are shown in Eq. \eqref{eq:cos}. 
\begin{equation}
    \label{eq:cvar}
    f(\tau;\phi) = \phi\tau,\quad \phi\in(0,1],\ \tau\sim U([0,1])    
\end{equation}
\begin{equation}
    \label{eq:cos}
    [\cos(\pi\cdot 0\cdot f(\tau;\phi)),\dots, \cos(\pi\cdot 63\cdot f(\tau;\phi))]
\end{equation}
During the training process, $\phi=1.0$, the number of samples in Eq. \eqref{eq:IQN loss} is $N=N'=8$, and the number of samples in Eq. \eqref{eq:approximate pi beta} is $K=32$.   
Given an observation, outputs of the DQN network are action values corresponding to actions, and those of the IQN network are sets of quantile points that reflect the return distributions of their respective actions.  

Both agents use $\epsilon$-greedy as their behavior policy.
The control time step of an action is 1.0 seconds. 
The exploration rate starts at 1.0, and decreases linearly to 0.05 during the initial 10\% of total training time steps, then
stays at 0.05 thereafter.
Other hyperparameters shared by both agents are shown in Table \ref{tab:hyperparameters}.

\begin{table}
    \centering
    \caption{\textbf{Hyperparameters} used by IQN and DQN during the learning process.}
    \begin{tabular}{|c|c|c|c|c|}
      \hline
      \multirow{2}{*}{Parameter} & Learning & Buffer & Mini-batch & Discount \\
      & rate & size & size & factor \\ \hline
      Value & $1\times 10^{-4}$ & $1\times 10^{6}$ & 32 & 0.99 \\ \hline
    \end{tabular}
    \vspace{-2mm}
    \label{tab:hyperparameters}
\end{table}

\subsection{Model Training}
\begin{figure}
    \centering
    \includegraphics[width=\linewidth]{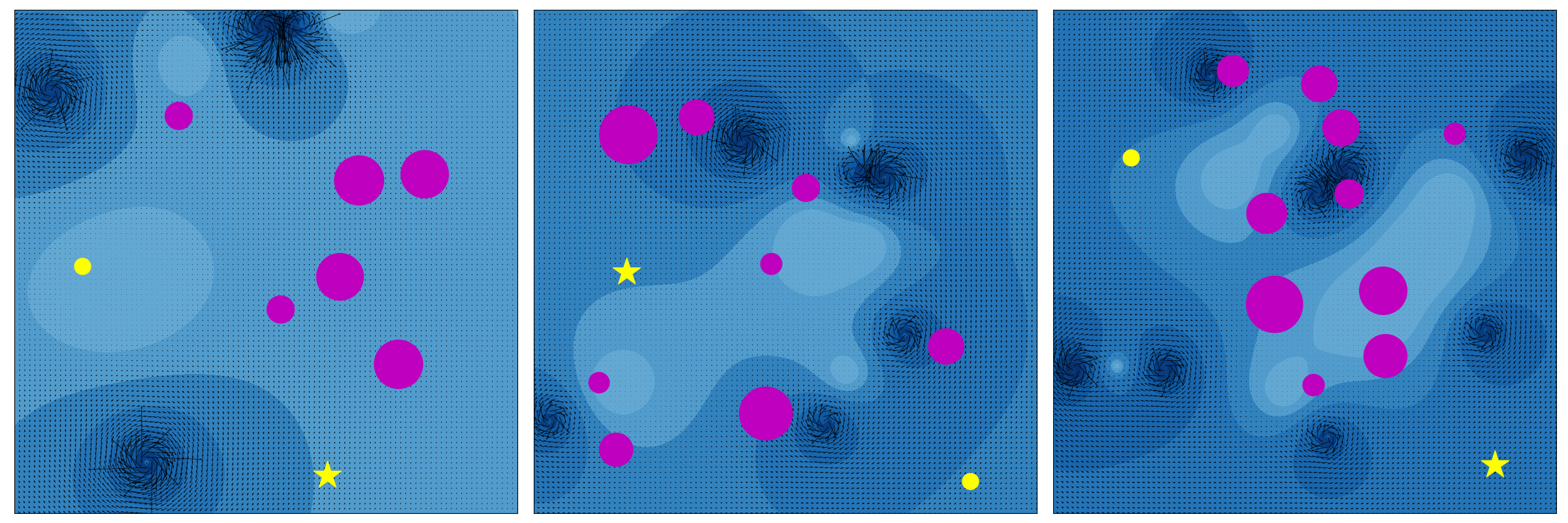}
    \caption{\textbf{Training environments.} Examples of training environments randomly generated for use during the 1st, 2nd and 3rd million steps (shown from left to right).}
    \vspace{-6mm}
    \label{fig:envs}
\end{figure}

As we focus on the local path planning problem, all simulation environments are of the size $50\text{m}\times50\text{m}$.
We employ the idea of curriculum training and gradually increase the level of difficulty in the training environment using the schedule shown in Table  \ref{tab:schedule}.
Example environments used during each phase are shown in Figure \ref{fig:envs}.

When initializing a training episode, vortices of random position, spinning direction, and strength, as well as circular obstacles of random position and size are generated in the training environment.
The linear velocity at the edge of vortex core, $v_{\text{edge}}$, and the radius of obstacle are sampled uniformly from $[5,10] \text{m/s}$ and $[1,3] \text{m}$ respectively.
The start and goal positions are randomly generated such that the distance between them is not smaller than the given threshold.
The initial robot pose and forward speed are also randomly generated. 
An episode is terminated when the robot collides with any obstacles, reaches the goal, or has moved for more than 1,000 steps, then a new episode is created.

\begin{table}
    \centering
    \caption{\textbf{Training Schedule}. Environment hyperparameters used in curriculum training.}
    \begin{tabular}{|c|c|c|c|}
         \hline
         Parameter & 1st million & 2nd million & 3rd million \\ \hline
        Number of vortices & 4 & 6 & 8 \\ \hline
        Number of obstacles & 6 & 8 & 10 \\ \hline
        Min distance between & \multirow{2}{*}{30.0} & \multirow{2}{*}{35.0} & \multirow{2}{*}{40.0} \\
        start and goal & & & \\ \hline
    \end{tabular}
    \label{tab:schedule}
\end{table}

\begin{figure}[h]
    \centering
    \includegraphics[width=\linewidth]{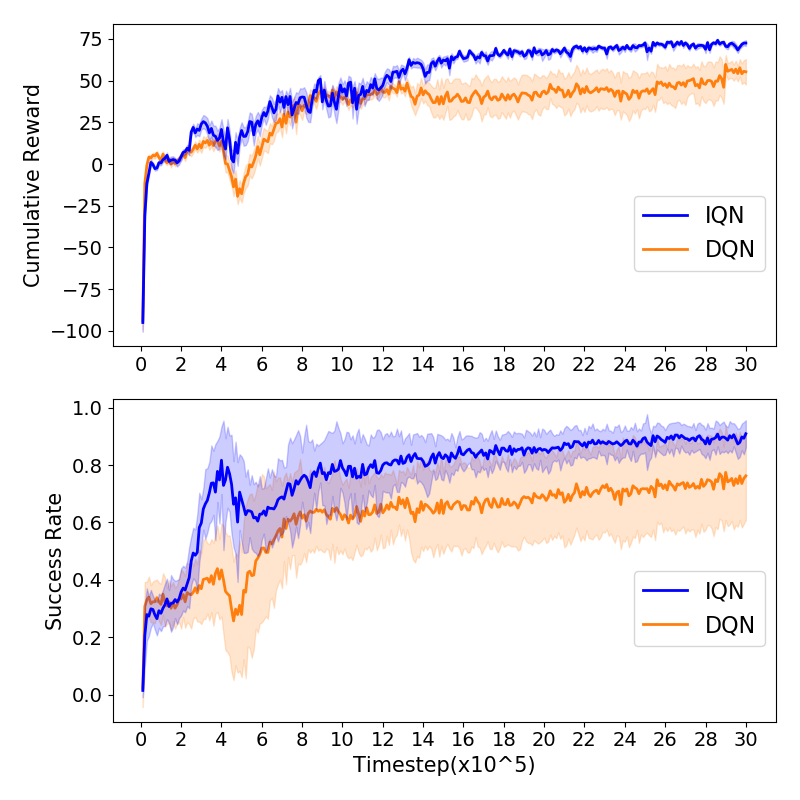}
    \caption{\textbf{Learning performances of IQN and DQN}. In each curve, the solid line and bandwidth represent the mean and standard error of evaluation results over thirty models trained with different seeds.}
    \label{fig:learning curves}
    \vspace{-1.5mm}
\end{figure}

To assess agents' performances during the training, thirty random environments are created in advance, which consist of three sets of ten environments generated according to parameters shown in each column in Table \ref{tab:schedule}, except the min distance between start and goal.
Every evaluation environment uses the same start and goal located respectively at the lower left and top right of the map. 
Every agent is evaluated on all thirty environments after every 10,000 steps of training. 
For both IQN and DQN, we train thirty models with different random seeds, and visualize the overall learning performance in Figure \ref{fig:learning curves}.

The training processes of all models were run on an Nvidia RTX 3090 GPU.
It can be seen that IQN exhibits a more stable learning performance and has higher scores in terms of cumulative reward and success rate.

\section{Experiments}
\label{sec:experiments}

To obtain a more comprehensive understanding of the performance of our trained agents, two sets of evaluation experiments using the number of vortexes and obstacles corresponding to the lowest and highest level of difficulty in Table \ref{tab:schedule} are performed, denoted as Test Case 1 and Test Case 2, and the results are summarized in Table \ref{tab:exp results}.
Each set of experiments uses 500 randomly generated environments, and similar to the evaluation environments during the training process, each environment has a fixed start and goal located respectively at the lower left and top right of the map.
The control time step of an action is set to 0.5 seconds.
A square boundary is set in each environment, and the robot moving outside the bounded area is considered an out of bounds failure.
The energy consumption is computed by summing up the magnitude of all action commands.  

IQN and DQN agents used in the evaluation experiments are randomly selected from trained models.  
We evaluate the IQN agent with fixed CVaR threshold values $\phi = \{0.25,0.50,0.75,1.0\}$ that represent different levels of sensitivity 
towards risk.
In addition, we also try an adaptive function of $\phi$ shown in Eq. \eqref{eq:adaptive cvar}.
\begin{equation}
    \label{eq:adaptive cvar}
    \phi = \left\{
            \begin{array}{lr}
                 \min(d(X,X_O))/d_0, &\text{if}\ \min(d(X,X_O))\leq d_0 \\
                 1.0, &\text{if}\ \min(d(X,X_O))> d_0
            \end{array}
            \right.
\end{equation}
$X$ and $X_O$ are positions of the robot and all obstacles.
Thus the adaptive framework leads to a greedy policy when no obstacles are detected, and a risk sensitive policy (with the level of sensitivity being proportional to the minimum distance to obstacles) if any obstacles exist. 

\subsection{Baseline approaches}

We also choose two classical local planning methods as baselines. 
The first one is an Artificial Potential Field (APF) method described in \cite{fan2020improved}.
An attractive potential field $U_{\text{att}}$ and a repulsive potential field $U_{\text{rep}}$ are constructed to generate forces that lead the robot to the goal and repel it from obstacles.
\begin{equation}
    U_{\text{att}}(X) = \frac{1}{2} k_{\text{att}} \cdot d^2(X,X_g) 
\end{equation}
\begin{equation}
    \begin{array}{c}
    U_{\text{rep}}(X) = \left\{
            \begin{array}{lr}
                 U_o(X), &\text{if}\ d(X,X_o)\leq d_0 \\
                 0, &\text{if}\ d(X,X_o)> d_0
            \end{array}
    \right. \\[10pt]
    \text{where } U_o(X) = \frac{1}{2}k_{\text{rep}}(\frac{1}{d(X,X_o)}-\frac{1}{d_0})^2d^n(X,X_g)
    \end{array}
\end{equation}
In the above equations, $X$, $X_g$ and $X_o$ are positions of the robot, goal and an obstacle respectively, and we use $k_{\text{att}}=50.0$, $k_{\text{rep}}=500.0$, and $n=2$.
The total force $F = -\nabla U_{\text{att}}(X)-\nabla U_{\text{rep}}(X)$.
Given an observation, each LiDAR reflection point is treated as an obstacle, and the repulsive force is the sum of contributions from all points.
To output an action decision that is compatible with the robot, we compute the difference in angle between the total force $F$ and the robot velocity, and map it to the angular velocity that causes the closest change in angle in one control step. 
We also project $F$ to the direction of robot velocity, scale it by $1/m$ with $m=500.0$, and map it to the closest linear acceleration. 

The second baseline is a Bug Algorithm (BA) method similar to VisBug \cite{lumelsky1988paradigm}.
It uses a simple navigation policy: when the robot hits an obstacle, it moves by following the boundary; as soon as the way to the goal is clear, the robot moves directly towards it.
Given an observation, the tangent vector of the obstacle surface is approximately computed from LiDAR reflections, and the robot steers itself and moves in the same direction while keeping a safe standoff distance of 5 meters.
Similarly, the angular velocity that minimizes the angle difference between desired steering direction and the robot velocity in one control step is chosen.
The linear acceleration is selected such that the robot maintains a low speed when moving along obstacles, and the max speed when moving towards the goal.

\subsection{Results}

It can be seen in Table \ref{tab:exp results} that all IQN agents except $\phi = 0.25$ achieve a higher success rate than other methods, with lower average time and energy consumption.
In addition, as the difficulty of environment increases, IQN agents show more robust performance in terms of success rate, with no significant increase in time and energy consumption.
It can be seen in Figure \ref{fig:demonstration} that the trajectories of APF and BA agents are highly affected by flow disturbances exerted by vortices.
Hence compared to DRL agents, APF and BA have clearly higher time and energy consumption, especially in Test Case 2.
The lower success rate in the case of $\phi = 0.25$ is related to the significantly higher out of bounds rate, indicating that the IQN agent is so conservative that it tries moving far enough to avoid any risks.
The adaptive IQN agent has a higher success rate than other IQN agents, with similar time and energy consumption.

\begin{table}[h!]
    \centering
    \caption{\textbf{Experimental Results}.
    Out of bounds rate is the ratio of the number of out of bounds failures to that of all episodes. 
    Average time and energy are computed using the data of successful episodes only.}
    \label{tab:exp results}
    \subcaption{\textbf{Test Case 1} (4 vortices and 6 obstacles)}
    \begin{tabular}{|>{\centering\arraybackslash}m{0.5cm}|>{\centering\arraybackslash}m{1.5cm}|>{\centering\arraybackslash}m{1.0cm}|>{\centering\arraybackslash}m{1.0cm}|>{\centering\arraybackslash}m{1.0cm}|>{\centering\arraybackslash}m{1.0cm}|}
        \hline
        \multicolumn{2}{|c|}{\multirow{3}{*}{Agent}} &
        \multirow{3}{*}{\shortstack{success \\ \\ rate}} & out of & \multirow{3}{*}{\shortstack{average \\ \\ time (s)}}  & \multirow{3}{*}{\shortstack{average \\ \\ energy}} \\ 
        \multicolumn{2}{|c|}{} &  & bounds rate &  &  \\ \hline
        \multirow{5}{*}{IQN} & adaptive & 0.95 & 0.01 & 35.32 & 85.34  \\ \cline{2-6}
        & $\phi$ = 0.25 & 0.87 & 0.11 & 36.80 & 86.46 \\ \cline{2-6}
        & $\phi$ = 0.50 & 0.94 & 0.02 & 35.35 & 83.62 \\ \cline{2-6}
        & $\phi$ = 0.75 & 0.94 & 0.02 & 35.43 & 83.14 \\ \cline{2-6}
        & $\phi$ = 1.0 & 0.94 & 0.01 & 34.73 & 82.80 \\
        \hline
        \multicolumn{2}{|c|}{DQN} & 0.88 & 0.02 & 35.58 & 106.16 \\ \hline
        \multicolumn{2}{|c|}{APF} & 0.90 & 0.01 & 44.58 & 124.47 \\ \hline
        \multicolumn{2}{|c|}{BA} & 0.66 & 0.00 & 42.31 & 104.35 \\ \hline
    \end{tabular}
    \newline
    \subcaption{\textbf{Test Case 2} (8 vortices and 10 obstacles)}
    \begin{tabular}{|>{\centering\arraybackslash}m{0.5cm}|>{\centering\arraybackslash}m{1.5cm}|>{\centering\arraybackslash}m{1cm}|>{\centering\arraybackslash}m{1cm}|>{\centering\arraybackslash}m{1cm}|>{\centering\arraybackslash}m{1cm}|}
        \hline
        \multicolumn{2}{|c|}{\multirow{3}{*}{Agent}} &
        \multirow{3}{*}{\shortstack{success \\ \\ rate}} & out of & \multirow{3}{*}{\shortstack{average \\ \\ time (s)}}  & \multirow{3}{*}{\shortstack{average \\ \\ energy}} \\ 
        \multicolumn{2}{|c|}{} &  & bounds rate &  &  \\ \hline
        \multirow{5}{*}{IQN} & adaptive & 0.81 & 0.04 & 36.76 & 93.36 \\ \cline{2-6}
        & $\phi$ = 0.25 & 0.60 & 0.29 & 39.38 & 97.78 \\ \cline{2-6}
        & $\phi$ = 0.50 & 0.76 & 0.09 & 37.39 & 92.85 \\ \cline{2-6}
        & $\phi$ = 0.75 & 0.78 & 0.07 & 35.90 & 88.69 \\ \cline{2-6}
        & $\phi$ = 1.0 & 0.76 & 0.03 & 35.42 & 87.94 \\
        \hline
        \multicolumn{2}{|c|}{DQN} & 0.61 & 0.04 & 34.67 & 106.19 \\ \hline
        \multicolumn{2}{|c|}{APF} & 0.58 & 0.07 & 53.72 & 165.03 \\ \hline
        \multicolumn{2}{|c|}{BA} & 0.37 & 0.01 & 49.73 & 133.15 \\ \hline
    \end{tabular}
    \vspace{-4mm}
\end{table}

\begin{figure*}[th]
    \centering
    \includegraphics[width=\textwidth]{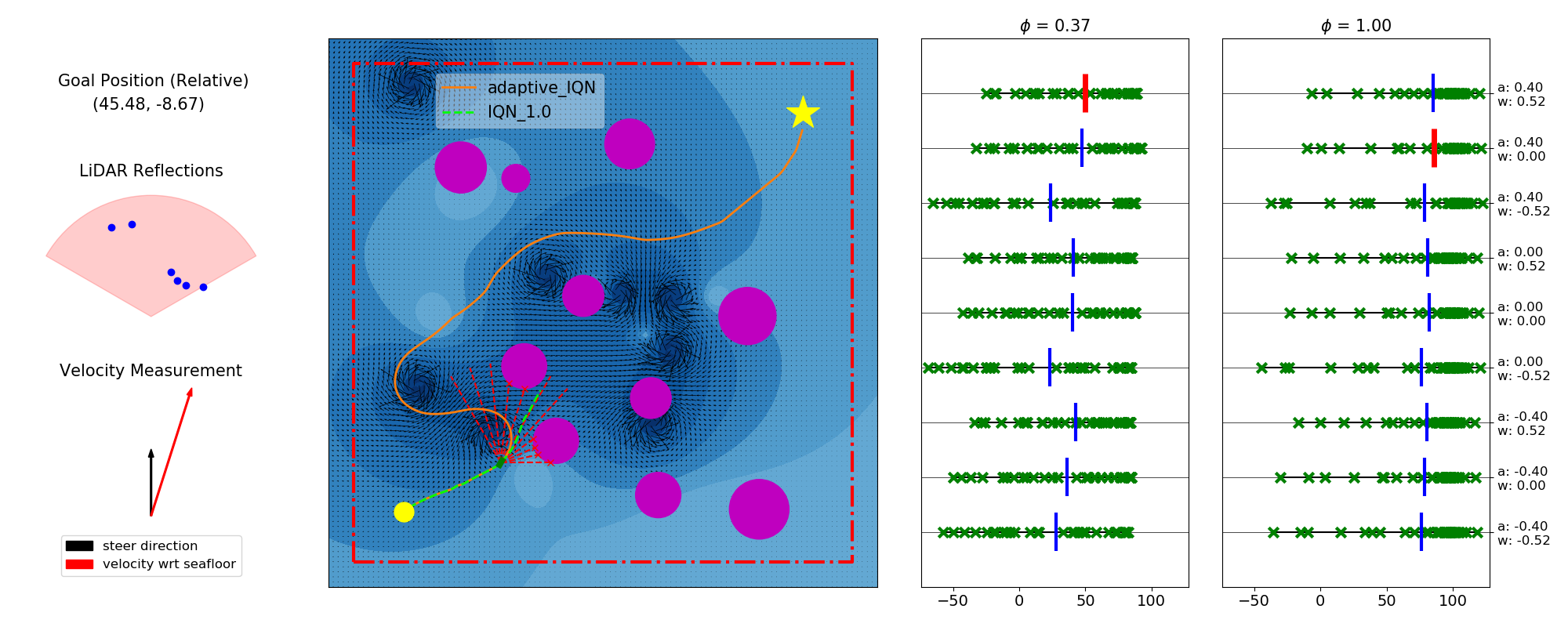}
    \caption{\textbf{An example state where adaptive IQN agent and greedy ($\phi=1.0$) IQN agent make different decisions}. Observation inputs are shown at left. 
    Trajectories of the two agents are shown in the map, where red dashed lines represent beams emitted by onboard LiDAR, and the red dashdot square is the boundary of the evaluation environment.
    In the distribution plots at right, green ``x" markers show return values corresponding to different quantiles, and vertical lines indicate the mean value (red lines mark action selections).}
    \vspace{-4mm}
    \label{fig:cvar_distributions}
\end{figure*}

Figure \ref{fig:cvar_distributions} visualizes a state where obstacles are detected and compares decisions made by the adaptive IQN agent and the greedy ($\phi = 1.0$) IQN agent. 
It can be seen from the observation plot that the goal direction is close to the steer direction as well as the velocity direction, and LiDAR reflections show a clearance between two sets of obstacles.
Hence sticking with the current direction may get to the goal quickly, which is also the action selected by the greedy IQN agent.
The adaptive IQN agent lowers the CVaR threshold to increase the risk sensitivity, and focuses on the worst part of the whole distribution.
In this case, turning left is a better choice as it has a higher expected return value in the tail distribution.
Additionally, the trajectories of the greedy and adaptive IQN agents overlap initially, since the latter one maintains $\phi = 1.0$ and efficiently navigates to the goal when no obstacles are detected.

The computation time per action decision for all methods is shown in Table \ref{tab:runtime}.
All experiments in this section were run on a AMD Ryzen threadripper 3970X CPU.
Compared to the classical methods, IQN and DQN models require more computation in one forward pass, but their mean runtimes per action are still below 0.4 milliseconds.
The maximum runtime per action of adaptive IQN is 13.12 milliseconds, which still allows a 75 Hz computation frequency.

\begin{table}[h]
    \centering
    \caption{\textbf{Runtime per action}. The mean and maximum runtime to compute an action decision.}
    \begin{tabular}{|>{\centering\arraybackslash}m{0.5cm}|>{\centering\arraybackslash}m{1.5cm}|>{\centering\arraybackslash}m{2cm}|>{\centering\arraybackslash}m{2cm}|}
        \hline
        \multicolumn{2}{|c|}{Agent} & mean (ms) & max (ms) \\ \hline
        \multirow{5}{*}{IQN} & adaptive & 0.31 & 13.12 \\ \cline{2-4}
        & $\phi$ = 0.25 & 0.28 & 2.11 \\ \cline{2-4}
        & $\phi$ = 0.50 & 0.28 & 0.87 \\ \cline{2-4}
        & $\phi$ = 0.75 & 0.28 & 1.29 \\ \cline{2-4}
        & $\phi$ = 1.0 & 0.28 & 4.35 \\
        \hline
        \multicolumn{2}{|c|}{DQN} & 0.19 & 4.48 \\ \hline
        \multicolumn{2}{|c|}{APF} & 0.068 & 0.22 \\ \hline
        \multicolumn{2}{|c|}{BA} & 0.049 & 1.81 \\ \hline
    \end{tabular}
    \label{tab:runtime}
    \vspace{-4mm}
\end{table}

\section{Conclusion}
\label{sec:conclusion}

In this paper, we propose an IQN based local path planner for  sensor-based Unmanned Surface Vehicle (USV) navigation in marine environments with no prior knowledge of the current flow field and obstacles.
Compared to a DQN based planner, the proposed method has a more stable learning performance and higher scores in the cumulative reward and success rate during the curriculum training process.   
Experimental results show that the proposed planner achieves superior performance in safety, time and energy consumption relative to other planners based on DQN, Artificial Potential Fields, and the Bug Algorithm. 
In future work, we hope to explore the multi-USV navigation problem with a similar approach, and the extent to which the policies learned in our simulator across different random obstacle and vortex fields can transfer to USV navigation with real robot hardware.

\section*{ACKNOWLEDGMENT}

This research was supported by the Office of Naval Research, grants N00014-20-1-2570 and  N00014-21-1-2161.

\bibliographystyle{IEEEtran}
\bibliography{main}

\end{document}